\documentclass{article}
\usepackage{ijcai24}

\usepackage[table,xcdraw]{xcolor}
\usepackage{times}
\usepackage{soul}
\usepackage{url}
\usepackage[hidelinks]{hyperref}
\usepackage[utf8]{inputenc}
\usepackage[small]{caption}
\usepackage{graphicx}
\usepackage{amsmath}
\usepackage{amsthm}
\usepackage{amsfonts}
\usepackage{booktabs}
\usepackage{algorithm}
\usepackage{algorithmic}
\usepackage[switch]{lineno}
\usepackage{enumitem}
\usepackage{multirow}
\usepackage{pifont}
\usepackage{wrapfig}
\usepackage[version=4]{mhchem}

\usepackage{float}
\usepackage{lipsum}
\usepackage{multicol}
\usepackage{siunitx}
\usepackage{tabularx,booktabs}
\usepackage{etoolbox}
\usepackage{newtxtext}
\usepackage{newtxmath}
\usepackage{booktabs}
\usepackage{threeparttable}

\usepackage{amsmath, amssymb, caption, subcaption, multirow, overpic, textpos}

\definecolor{LightCyan}{rgb}{0.88,1,1}
\definecolor{LightRed}{rgb}{1,0.88,1}
\definecolor{LightYellow}{rgb}{1,1,0.88}
\definecolor{Grey}{rgb}{0.75,0.75,0.75}
\definecolor{DarkGrey}{rgb}{0.55,0.55,0.55}
\definecolor{DarkGreen}{rgb}{0,0.65,0}

\newlength\savewidth\newcommand\shline{\noalign{\global\savewidth\arrayrulewidth
  \global\arrayrulewidth 1pt}\hline\noalign{\global\arrayrulewidth\savewidth}}
\newcommand{\tablestyle}[2]{\setlength{\tabcolsep}{#1}\renewcommand{\arraystretch}{#2}\centering\footnotesize}
\definecolor{baselinecolor}{gray}{.9}
\newcommand{\baseline}[1]{\cellcolor{baselinecolor}{#1}}

\newcolumntype{x}[1]{>{\centering\arraybackslash}p{#1pt}}
\newcolumntype{y}[1]{>{\raggedright\arraybackslash}p{#1pt}}
\newcolumntype{z}[1]{>{\raggedleft\arraybackslash}p{#1pt}}

\newcommand{\tianhong}[1]{\textcolor{blue}{\ignorespaces}}
\newcommand{\huiwen}[1]{\textcolor{red}{\ignorespaces}}
\newcommand{\han}[1]{\textcolor{cyan}{\ignorespaces}}
\newcommand{\dilip}[1]{\textcolor{orange}{\ignorespaces}}
\newcommand{\shlok}[1]{\textcolor{green}{\ignorespaces}}

\colorlet{darkgreen}{green!65!black}
\colorlet{darkblue}{blue!75!black}
\colorlet{darkred}{red!80!black}
\definecolor{lightblue}{HTML}{0071bc}
\definecolor{lightgreen}{HTML}{39b54a}

\usepackage{listings} % for code listing
\definecolor{codeblue}{rgb}{0.28,0.24,0.55}
\definecolor{codeorange}{rgb}{0.78,0.41,0.08}

\lstdefinestyle{mystyle}{
    basicstyle=\bfseries\footnotesize,
    commentstyle=\color{codeblue},
    keywordstyle=\color{codeorange},
    breakatwhitespace=false,         
    breaklines=true,                 
    captionpos=b,                    
    keepspaces=true,                 
    numbers=none,                    
    numbersep=5pt,                  
    showspaces=false,                
    showstringspaces=false,
    showtabs=false,                  
    tabsize=2,
    frame=tb, % 在顶部和底部添加横线
    framerule=0.4pt, % 横线的粗细
    framesep=3pt, % 文本与框线的间距
    lineskip=3pt
}

\title{Where to Mask: Structure-Guided Masking for Graph Masked Autoencoders}

\author{
Chuang Liu$^1$\thanks{Equal Contribution}\and
Yuyao Wang$^{1*}$\and
Yibing Zhan$^2$\and
Xueqi Ma$^3$\and \\
Dapeng Tao$^4$\and
Jia Wu$^5$\and
Wenbin Hu$^1$\footnote{Corresponding Author} 
\affiliations
$^1$School of Computer Science, Wuhan University, Wuhan, China\\
$^2$JD Explore Academy, JD.com, China \\
$^3$School of Computing and Information Systems, The University of Melbourne, Melbourne, Australia \\
$^4$School of Computer Science, Yunnan University, Kunming, China \\
$^5$School of Computing, Macquarie University, Sydney, Australia
\emails
\{chuangliu, wyy0224, hwb\}@whu.edu.cn,
zhanyibing@jd.com,
xueqim@student.unimelb.edu.au,
dptao@ynu.edu.cn,
jia.wu@mq.edu.au
}

\begin{document}

\maketitle

\begin{abstract}
   Graph masked autoencoders (GMAE) have emerged as a significant advancement in self-supervised pre-training for graph-structured data. Previous GMAE models primarily utilize a straightforward random masking strategy for nodes or edges during training. However, this strategy fails to consider the varying significance of different nodes within the graph structure. In this paper, we investigate the potential of leveraging the graph's structural composition as a fundamental and unique prior in the masked pre-training process.  To this end, we introduce a novel structure-guided masking strategy (\textit{i.e.}, StructMAE), designed to refine the existing GMAE models. StructMAE involves two steps: \textbf{1)} Structure-based Scoring: Each node is evaluated and assigned a score reflecting its structural significance.  Two distinct types of scoring manners are proposed: predefined and learnable scoring. \textbf{2)} Structure-guided Masking: With the obtained assessment scores, we develop an easy-to-hard masking strategy that gradually increases the structural awareness of the self-supervised reconstruction task. Specifically, the strategy begins with random masking and progresses to masking structure-informative nodes based on the assessment scores. This design gradually and effectively guides the model in learning graph structural information. Furthermore, extensive experiments consistently demonstrate that our StructMAE method outperforms existing state-of-the-art GMAE models in both unsupervised and transfer learning tasks. Codes are available at \url{https://github.com/LiuChuang0059/StructMAE}.

\end{abstract}

% a predefined method based on established structural criteria, and a learnable method that adapts dynamically to the learning process.

% we aim to improve GMAE by introducing a structure-guided masking strategy (\textit{i.e.}, StructMAE). 

% StructMAE's main principle is that leveraging the graph's structural composition as a crucial and unique prior during masked pre-training can significantly enhance the model's learning capabilities.

% Most previous works mask patches of the image randomly, which underutilizes the semantic information that is beneficial to visual representation learning. 

%   However, this straightforward random masking approach may result in two extremes: either masking predominantly simple, non-critical nodes, impeding the model's learning of valuable knowledge, or masking crucial nodes, making learning excessively challenging.  

\section{Introduction}
\label{sec:introduction}

\begin{figure}[!t] % !ht% \setlength{\abovecaptionskip}{-0.1cm}   %调整图片标题与图距离
\begin{center}
\includegraphics[width=1.0\linewidth]{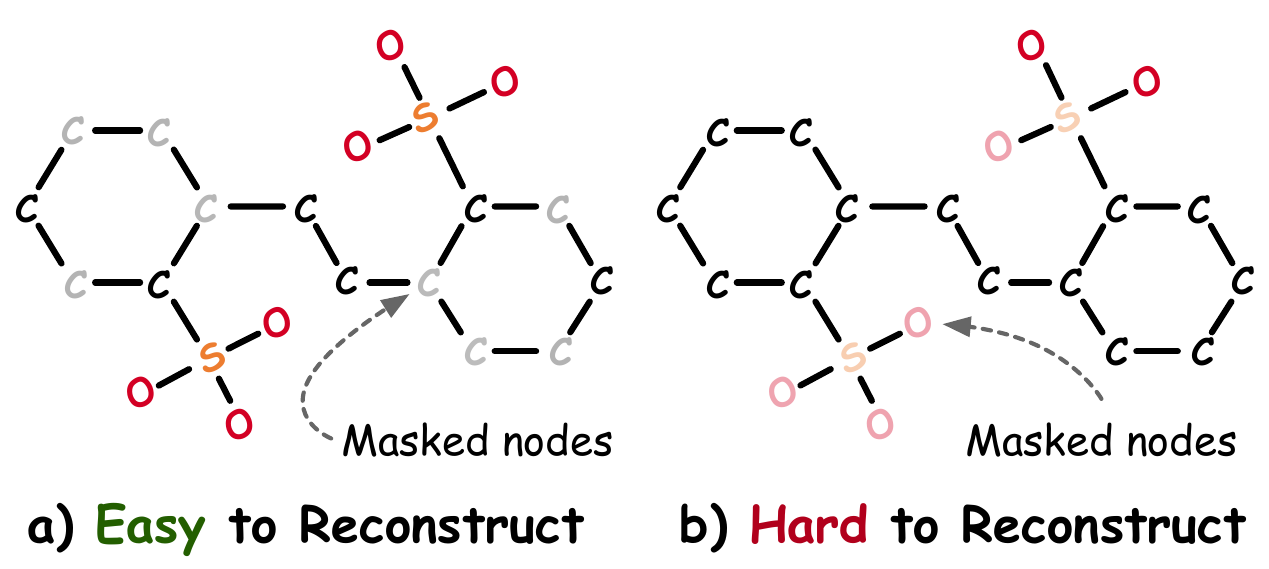}
\end{center}
\caption{Two primary examples that underscore the potential suboptimal nature of the \textit{random masking} strategy in GMAE.  \textbf{a)} Masked nodes are too simplistic to predict (\textit{i.e.}, \ce{C}), hindering the acquisition of valuable knowledge. \textbf{b)} Masking a large number of informative chemical nodes (\textit{i.e.}, \ce{SO3}) makes the model fail to perceive the structural information in graphs.}
\label{fig:motiv}
\end{figure}

In domains, such as academic, social, and biological networks, graph-structured data often lacks labels. This problem is especially acute in biochemical graphs due to the expense of wet-lab experiments. To address this challenge, many techniques have been developed to fully exploit the existing massive amounts of unlabeled data, aiming to enhance graph model training. Among these approaches, self-supervised graph pre-training (SSGP) methods are prominent due to their effectiveness, attracting significant interest in academic and industrial realms~\cite{survey-pretraining}. 

Currently, SSGP methods are categorized into two primary streams:  \textbf{1) Contrastive} methods, such as GraphCL~\cite{graphCL} and SimGRACE~\cite{simGrace}, which utilize contrastive learning principles to reveal the intrinsic structure and interconnections within graph data. \textbf{2) Generative} methods, including GraphMAE~\cite{graphmae} and MaskGAE~\cite{maskgae}. These methods focus on learning node representations through a reconstruction objective. Moreover, generative methods have proved to be simpler and more effective than contrastive approaches that require carefully designed augmentation and sampling strategies. The efficacy of generative methods is further underscored by the enormous successes of models such as BERT and ChatGPT in Natural Language Processing (NLP)~\cite{bert} and MAE in Computer Vision (CV)~\cite{mae}. These successes highlight the significant potential of generative approaches across various domains. Accordingly, this paper explores the capabilities of generative methods, specifically graph masked autoencoders (GMAE), in graph learning tasks and recognizes their potential as evidenced in other fields.

% Additionally, in the realm of graph learning, recent generative models, such as graph masked autoencoders (GMAE), have surpassed earlier contrastive pre-training models.

% Though random masking could serve as a strong strategy for MIM, several obvious problems are required to resolve.

GMAE fundamentally involves randomly masking a proportion of input data (\textit{i.e.}, nodes or edges) and leveraging the reconstruction of the removed contents to guide the representation learning. Despite GMAE's promising results, its random masking approach, which assigns equal probability to all nodes in a graph - a universally adopted strategy, presents a suboptimal strategy. Specifically, the masked nodes are sometimes overly simplistic to predict (\textit{i.e.}, the atom \ce{C} in Figure~\ref{fig:motiv} (a)) with only neighborhood information. In such cases, the model's pre-training phase may not be sufficiently informative, thereby hindering the acquisition of valuable knowledge. However, if we mask a large number of key informative nodes (\textit{i.e.}, \ce{SO3} in Figure~\ref{fig:motiv} (b)), the model may fail to perceive the graph’s overall structural information. In summary, the indiscriminate nature of random masking, which fails to distinguish nodes of varying informational values, potentially leads to low data efficiency and compromises the quality of the learned graph representations.  Therefore, this raises the question: \textit{is there a more effective  masking strategy than random sampling for enhancing GMAE's pre-training process? }

% Based on the above discussion, we design a preliminary experiments, as illustrated in Figure.  From the results, we find that both scenarios result in the degeneration of the learned representations.

% In our preliminary experiments, we find that both scenarios result in the degenera- tion of the learned visual representations, as illustrated in Section 2.

% random masking is prone to make at- tention dispersed on the whole image and not sufficiently focused on the object.

% we propose to create a curriculum by generating masks of increasing difficulty during training.

This paper answers the aforementioned open question by introducing StructMAE, which features a novel structure-guided masking strategy designed to enhance GMAE's pre-training process. The key insight of our method is to inject prior graph structure knowledge into the masking process to guide model learning. Specifically, StructMAE comprises two principal components: \textbf{1) Structure-based Scoring:} We recognize that the node reconstruction complexity is inherently linked to its structural significance within the graph.  Therefore, we derive a scoring method to assess the node significance, distinguishing between informative and less-informative nodes based on structural considerations. In addition, two scoring method variants are proposed: predefined and learnable, which are discussed in detail in Section~\ref{sec:method}. \textbf{2) Structure-guided Masking:} With node significance scores established, we propose an easy-to-hard masking strategy that gradually increases the difficulty of the self-supervised reconstruction task. This approach commences with the masking of less-informative nodes, progressively shifting towards masking more informative nodes as the model's learning progresses. This strategic progression in masking difficulty is designed to enable the model to gradually and effectively assimilate the graph's structural information.

To evaluate the effectiveness of the StructMAE model, we conduct comprehensive experiments on a range of widely-used datasets, notably the large-scale Open Graph Benchmark (OGB)~\cite{ogb-dataset},  covering two graph learning tasks: unsupervised and transfer learning. The experimental results consistently demonstrate that StructMAE's performance surpasses that of existing state-of-the-art models in contrastive and generative pre-training domains. This exceptional performance demonstrates our structure-guided masking approach's advantages over conventional random masking methods. The principal contributions are summarized as follows:

\begin{enumerate}[leftmargin=12pt]
  \item  We introduce StructMAE, a novel node masking strategy tailored for GMAE. This strategy utilizes the structural information inherent in graphs to gradually and effectively direct the masking process, significantly enhancing the model's capability for representation learning.

  % \item We develop two distinct scoring methods from the graph structure perspective. Based on the obtained node scores, an easy-to-hard progressive masking strategy is designed to enhance the learning process of the pre-training model in gradually and effectively absorbing graph knowledge.

  \item We evaluate StructMAE through extensive experiments, comparing its performance with generative and contrastive baselines across two graph tasks on various real-world graph datasets, including the OGB dataset. The experimental results consistently validate StructMAE's effectiveness.
\end{enumerate}

% 1. We explore the masking strategies in MIM and demon- strate that the previous random sampling method is sub- optimal. We find that the pre-training results can be sig- nificantly improved by slightly raising the masking rates of the informative foreground image patches.

% 2. It is worth noting that the proposed  is a plug-and- play module for GMAE frameworks and could be readily deployed into GMAE methods, such as GraphMAE and MaskGAE.

% As a plug-and-play module, 

\section{Related Work}
\label{sec:related work}

\paragraph{Self-supervised Graph Pre-training.} 
Inspired by the success of pre-trained language models like BERT~\cite{bert}, T5~\cite{t5}, and ChatGPT~\cite{gpt}, numerous efforts have been directed towards SSGP. Based on model architectures and objective designs, SSGP is naturally divided into contrastive and generative methods. First, contrastive self-supervised learning has dominated graph representation learning in the past two years. Its success is largely due to elaborate data augmentation designs, negative sampling, and contrastive loss. For instance, DGI~\cite{dgi} and InfoGraph~\cite{infoGraph:}, based on mutual information, leverage corruptions to construct negative pairs. Similarly, models such as SimGRACE~\cite{simGrace} and GraphCL~\cite{graphCL} utilize in-batch negatives.  Differing from previous methods that execute augmentation on graphs, COSTA~\cite{costa} implements augmentation within the embedding space, aiding in the mitigation of sampling bias. Moreover, POT~\cite{pot}  advances graph contrastive learning (GCL) training through the employment of a node compactness metric, assessing adherence to the GCL principle. Second, generative self-supervised learning focuses on recovering missing parts of the input data. For example, GAE~\cite{gae} is a conventional method that reconstructs the adjacency matrix. Multiple GAE variants utilize graph reconstruction to pre-train Graph Neural Networks (GNNs), including  ARVGA~\cite{ARgae} and SIGVAE~\cite{siggae}. Recently, a paradigm shift towards GMAE has shown promising results in various tasks~\cite{mgap}. A detailed introduction will be provided in the following section.

% Compared to contrastive methods, GAEs are generally very simple to implement and easy to combine with existing frameworks, since they naturally leverage graph reconstruction as pretext tasks without need of augmentations for view generations.

\begin{figure*}[!t] % !htb
\begin{center}
\includegraphics[width=1.0\linewidth]{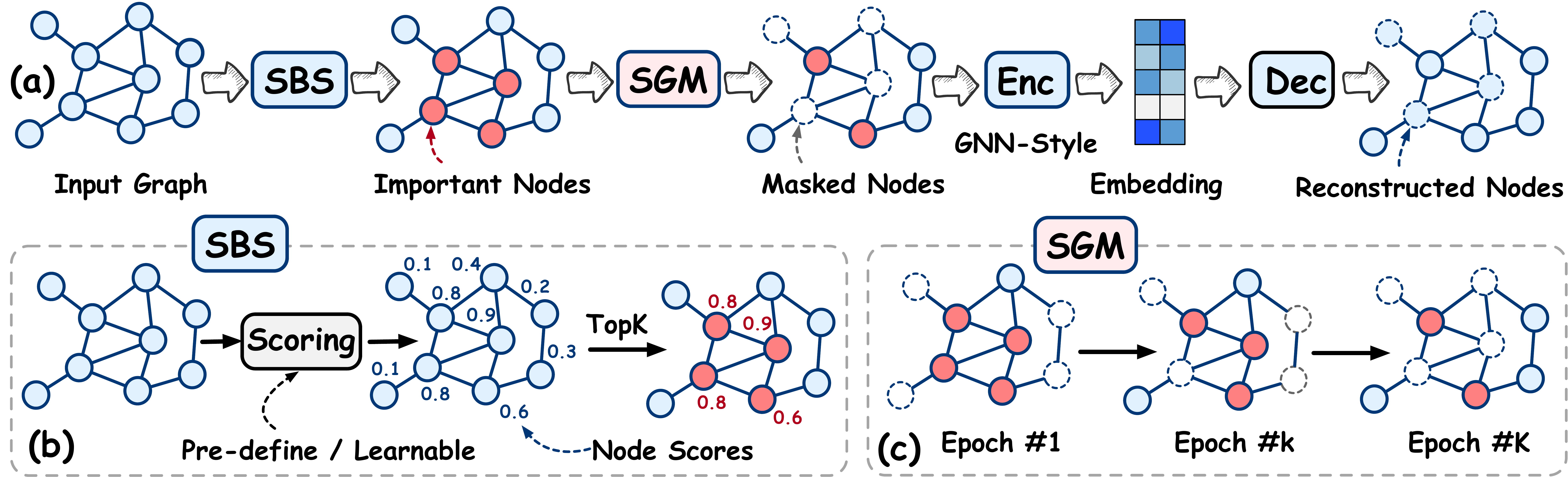}
\end{center}
\caption{Overview of the proposed model. \textbf{(a)} The overall architecture of the proposed StructMAE. \textbf{(b)} SBS: It evaluates node importance based on the graph's structural information. This evaluation can be conducted using either a predefined or learnable approach.  \textbf{(c)} SGM: It progressively increases the masking probability of important nodes as the training epochs advance.}
\label{fig:model}
\end{figure*}

\paragraph{Graph Masked Autoencoders.} 
GMAE mainly involves reconstructing the contents (\textit{e.g.}, nodes and edges) that are randomly masked from the input using autoencoder architecture. A notable example of GMAE is GraphMAE~\cite{graphmae}, which reconstructs randomly masked input node features with several innovative designs, including re-mask decoding and scaled cosine error. In addition, MaskGAE~\cite{maskgae}, S2GAE~\cite{s2gae}, and GiGaMAE~\cite{gigamae} jointly reconstruct the masked edges and node degrees. SimSGT~\cite{simsgt}, and GCMAE~\cite{GCMAE} combine contrastive learning with GMAE, whereas RARE~\cite{rare} employs self-distillation to enhance GMAE's performance. Unlike the above mentioned GMAE methods, which primarily use message-passing GNNs as backbone models, GMAE-GT~\cite{gmae-gt} utilizes a graph transformer~\cite{graphormer-v1,gapformer} as its encoder backbone. There are also several GMAE applications, including heterogeneous graph representation learning~\cite{hgmae}, protein surface prediction~\cite{protein-mae}, and action recognition~\cite{skeletonmae}. However, all the aforementioned methods employ a random masking method when masking graph contents and thus overlook the importance of GMAE mask strategies, potentially inhibiting the model's capabilities. The work most closely related to this study is MoAma~\cite{moama}, which similarly focuses on designing superior mask strategies. However, this approach uses motifs which rely on domain knowledge and manual motif pre-definition, limiting its generalizability across various domains.

% Subsequently, GraphMAE2 extends GraphMAE to large-scale graphs using a local clustering method and introduces multi-view re-mask decoding.
% Masked language modeling (MLM) is the first successful application of masked autocoding in natural language processing. Its working principle is similar to the cloze test in English. Recently, masked image modeling (MIM) follows a similar principle by masking redundant pixel blocks and predicting them for learning. Although masked autocoding technology is very popular in language and visual research, it is relatively less studied in the graph domain.~\protect\cite{maskgae} 

% However, the above two GTs neglect the important structural information in a graph. Therefore, Ying \textit{et al.}~\shortcite{graphormer-v1} encoded the structural information of a graph, including centrality, spatial, and edge informations into GTs.

\section{Preliminaries}
\label{sec:preliminary}

\paragraph{Notations.}  A graph $\mathcal{G} =(\mathcal{V}, \mathcal{E}) $  can be represented by an adjacency matrix $\mathbf{A} \in \{0, 1\}^{ n \times n}$ and a node feature matrix $\mathbf{X} \in \mathbb{R}^{ n \times d}$, where $\mathcal{V}$ denotes the node sets, $\mathcal{E}$ denotes the edge sets,  $n$ is the number of nodes, $d$ is the dimension of the node features, and  $\mathbf{A}[i, j]=1$ if there exists an edge between nodes $v_{i}$ and $v_{j}$, otherwise, $\mathbf{A}[i, j]=0$.

\paragraph{Graph Masked Autoencoders.}  GMAE operates as a self-supervised pre-training framework, which focuses on recovering masked node features or edges based on the representations of unmasked nodes. To illustrate this method, we focus on the reconstruction of node features as an example. GMAE comprises two essential components: an encoder ($f_{E}(\cdot)$) and a decoder ($f_{D}(\cdot)$). The encoder maps each unmasked node $v \in \mathcal{V}_{\text{unmask}}$ to a $d$-dimensional vector $\mathbf{h}_v \in \mathbb{R}^d$, with $\mathcal{V}_{\text{unmask}}$ representing the set of unmasked nodes, while the decoder reconstructs the masked node features from these vectors. The entire process can be formally represented as:
\begin{equation}
\mathbf{H}_{\text{unmask}}=f_E(\mathbf{A}, \mathbf{X}_{\text{unmask}}); \quad \mathbf{X}^{\prime}=f_D(\mathbf{A}, \mathbf{H}_{\text{unmask}}),
\end{equation}
where $\mathbf{X}_{\text{unmask}}$ and $\mathbf{H}_{\text{unmask}}$ denote the features and embeddings of unmasked nodes, respectively, and  $\mathbf{X}^{\prime}$ represents the reconstructed features of all the nodes. Then, GMAE optimizes the model by minimizing the discrepancy between the reconstructed representation of masked nodes, $\mathbf{X}^{\prime}_{\text{mask}} \subset \mathbf{X}^{\prime}$,  and their original features, $\mathbf{X}_{\text{mask}} \subset \mathbf{X}$.

\section{Methodology}
\label{sec:method}

% \subsection{Proposed Method: StructMAE}
% \label{sec:propose-method}

This section presents the model architecture of StructMAE in detail. First, we provide a detailed introduction to the masking module in GMAE (\textbf{$\S$\ref{sec:intro-masking}}). Subsequently, a comprehensive exploration (\textbf{$\S$\ref{sec:rethink-mask}}) of the masking strategy is provided. Then, the details of the proposed StrucMAE are elucidated  (\textbf{$\S$\ref{sec:design-mask}}). Finally, the overall StructMAE architecture, encompassing training and inference details, is expounded  (\textbf{$\S$\ref{sec:overall-arch}}).

% Next, the training and inference procedures of StructMAE are detailed. Finally, an in-depth discussion on StructMAE's connection with contrastive learning is presented.

\subsection{Introducing the GMAE Masking Module} 
\label{sec:intro-masking}

Prior to processing graph data within the GMAE encoder, a subset of nodes $\mathcal{V}_{\text{mask}} \subset \mathcal{V}$ is sampled for masking. According to the methodology described in~\cite{graphmae}, the features of these sampled nodes are replaced with a learnable vector $\boldsymbol{x}_{[\text{M}]} \in \mathbb{R}^d$. Accordingly, for a node $v_i$ within the masked node subset $\mathcal{V}_{\text{mask}}$, its feature $\widetilde{\mathbf{x}}_i$ in the altered feature matrix $\widetilde{\mathbf{X}}$ is defined as follows:
\begin{equation}
\widetilde{\mathbf{x}}_i= \begin{cases}\mathbf{x}_{[\text{M}]} & v_i \in \mathcal{V}_{\text{mask}} \\ \mathbf{x}_i & v_i \notin \mathcal{V}_{\text{mask}}\end{cases}.
\end{equation}
Regarding the approach for selecting nodes to mask, most existing works~\cite{graphmae,maskgae} utilize a random masking strategy. This method assigns an equal masking probability of masking to each node within a graph. A more detailed exploration of the random masking strategy will be presented in the following section (\textbf{$\S$\ref{sec:rethink-mask}}).

% The objective of StructMAE is to reconstruct the masked features of nodes in $\widetilde{V}$ by utilizing the partially observed node signals $\widetilde{X}$.

\subsection{Reconsidering the Random Masking Strategy}
\label{sec:rethink-mask}
In GMAE context, the masking strategy is crucial, as it significantly influences the type of information that the model learns. In previous studies~\cite{graphmae},
nodes within a graph are masked randomly, each with an equal probability. However, this strategy overlooks the varying structural information of different nodes, which has been shown to be crucial in graph learning tasks. Therefore, we conduct a preliminary experiment to explore whether the incorporation of structure priors could augment pre-training efficacy.

\begin{figure}[!t] % !htb
\begin{center}
\includegraphics[width=0.95\linewidth]{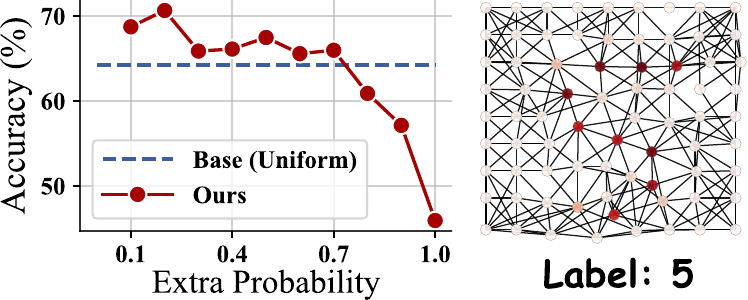}
\end{center}
\caption{Effects of raising the masking probability on nodes with structural information (dark red nodes in the right part). The blue dashed line illustrates the results under the random masking strategy.
}
\label{fig:toy}
\end{figure}

In the experiment, models are pre-trained on the MNIST dataset and evaluated in unsupervised settings. Nodes forming numerical values are identified as those possessing rich structural information. Subsequently, the masking probability for these structurally informative nodes is manually increased. The results, presented on the left side of Figure~\ref{fig:toy}, reveal the impact of this modified masking strategy on unsupervised accuracy. The results indicate that: \textbf{1)} A marginal increase in the masking probability for nodes with rich structural information enhances the model's pre-training learning. Furthermore, up to a probability threshold of 0.2, a corresponding gradual increase in the model's accuracy is observed. \textbf{2)} Conversely, excessively raising the masking probability of these structurally significant nodes detrimentally affects the model's training. These findings corroborate our initial discussion and highlight the importance of proposing an effective method to integrate structural information into the masking process.

% Consequently, it may lead to suboptimal learning outcomes, as it fails to differentiate between nodes that are more or less informative and challenging to reconstruct. 

% Most previous works~\cite{graphmae,maskgae} relied on an uncontrolled random mask on node features, such as random feature masking (Figure~\ref{fig:model} (a)), potentially leading to an inadequate training or learning challenge for the model.

\subsection{The Proposed Masking Strategy}
\label{sec:design-mask}
Inspired by the preceding discussion, we introduce an innovative structure-guided masking approach for GMAE, named StructMAE (as depicted in Figure~\ref{fig:model} (b)). StructMAE involves integrating the graph's structural knowledge into the masking process, thereby directing the model's learning trajectory more effectively. StructMAE is composed of two primary elements:  \textbf{Structure-based Scoring (SBS)} and \textbf{Structure-guided Masking (SGM)}.  

% The operational framework of StructMAE can be formally represented as follows:
% \begin{equation}
% \mathbf{x}_{\text{mask}}=\operatorname{SGM}\left( \operatorname{SBS}\left(\left\{\mathbf{x}_i: v_i \in \mathcal{V}\right\}\right)\right).
% \end{equation}

\subsubsection{Structure-based Scoring}
The SBS evaluates the significance of each node based on its structural role within the graph. This evaluation facilitates the identification of nodes that are pivotal for the model to learn, allowing for a more targeted masking approach. To determine the importance of nodes, we introduce two distinct methodologies: the pre-defined and learnable methods.

\paragraph{Pre-defined Structure-based Scoring.} The predefined method involves using a set of predetermined criteria, based on the known structural information, to evaluate node importance. Specifically, the computation of importance score $\mathbf{S} \in \mathbb{R}^n$ is achieved using the PageRank algorithm~\cite{pagerank}, a well-established technique for evaluating node significance based on graph structures. Thus, the importance score of node $v_i$ is defined as:
\begin{equation}
s_i= \frac{1-e}{n} + e \sum_{j \in \mathcal{N}_i} \frac{s_j}{L_j},
\end{equation}
where $e$ denotes the damping factor, $L_i$ represents the degree of node $v_i$, and $\mathcal{N}_i$ denotes the set of neighboring nodes of node $v_i$. In addition to PageRank, other prevalent metrics for assessing node importance comprise degree, closeness centrality, and betweenness centrality. Each of these methods is based on different underlying principles, offering diverse perspectives on a node’s role and influence within a graph.  Although these methods constitute more intricate ways of evaluating node importance, our empirical findings suggest that PageRank serves as a straightforward and effective measure. A detailed discussion and comparison of these methods is presented in the following section.

\paragraph{Learnable Structure-based Scoring.}  In contrast to the predefined method, the learnable approach dynamically assesses node significance based on the evolving state of the graph during the learning process. Specifically, it integrates the formulation of the assessment metric with masked graph modeling, enabling end-to-end learning of this metric. To accomplish this, we employ a lightweight scoring network, denoted as $f_S(\cdot)$, which assesses the importance of each node $v_i$. The scoring network's design is similar to a GNN-style layer, and it effectively captures graph structural information. Thus, the importance score $s_i \in \mathbf{S}$ for node $v_i$ is calculated as follows:
\begin{equation}
s_i=\operatorname{Sigmoid}\left( f_S\left(\mathbf{x}_i, \mathbf{A}\right)\right), \quad i=1, \ldots, n.
\end{equation}
A higher score $s_i$ indicates greater importance of the corresponding node $v_i$. Following the scoring, node features are sorted in descending order based on these scores. The ordered node features and their respective scores are represented as  $\left\{\mathbf{x}_i^{\prime}\right\}$ and $\left\{s_i^{\prime}\right\}$, respectively, where $i=1, \ldots, n$. To facilitate learning the scoring network $f_S(\cdot)$, the predicted scores are multiplied by the node features to serve as modulating factors. This operation is formally expressed as:
\begin{equation}
\begin{gathered}
\hat{\mathbf{X}}=\left\{\mathbf{x}_i^{\prime \prime} \mid \boldsymbol{x}_i^{\prime \prime}=\boldsymbol{x}_i^{\prime} * s_i^{\prime}\right\}, \quad i=1, \ldots, n,  
\end{gathered}
\end{equation}
where $\hat{\mathbf{X}}$ denotes the set of node features after scoring. This mechanism ensures that the scoring network is continually updated and refined throughout the model's training process.

In this section, we explore the two distinct SBS methods proposed for StructMAE. First, the pre-defined SBS method offers simplicity and is particularly effective when the structural characteristics are well-understood and can be explicitly defined beforehand. Conversely,  the learnable SBS is more adaptable and can cater to complex and variable graph structures, making it suitable for scenarios where the node significance cannot be easily predetermined.

\subsubsection{Structure-guided Masking}
The SGM component utilizes the scores generated by the SBS to guide its masking decisions. It selectively and progressively masks nodes in an easy-to-hard manner, thereby enhancing the model's capacity to effectively learn and represent graph structures.  Specifically, in the initial training stage, a subset of easy nodes with lower scores is masked, making it easier for the model to predict them using basic neighboring information. As training progresses, the masking strategy evolves to encompass more challenging nodes. This enables the model to capture intricate structural information and, consequently, enhances its learning capabilities.

This masking strategy relies on the importance scoring matrix $\mathbf{S}$. It is hypothesized that nodes with higher $\mathbf{S}$ scores are more informative and significant. Consequently, we gradually increase the masking probabilities for these high-scoring nodes during masked graph modeling. To implement this, we rank the nodes based on their scores $\mathbf{S}=\left\{s_1, s_2, \cdots, s_n\right\}$ and identify the top $K$ indices that constitute the set $\mathcal{Y}$. The masking probability for each node is then determined as:
\begin{equation}
\gamma_i=\epsilon+ \begin{cases}\beta & v_i \in \mathcal{Y} \\ 0 & v_i \notin \mathcal{Y}\end{cases},
\end{equation}
where $\epsilon$ denotes random noise drawn from a uniform distribution ($\epsilon \sim U(0,1)$) and $\beta$ represents the increased probability assigned to nodes with higher scores. In this instance, nodes in the set $\mathcal{Y}$ are considered more informative, and consequently, the model is anticipated to prioritize these nodes. The number of masked nodes, denoted by $K$, is dynamically adjusted throughout the training process. Initially set at zero, $K$ progressively increases with epoch according to the following formula:
\begin{equation}
K(t)=p \cdot n \cdot  \sqrt{t/T}, 
\end{equation}
where $K(t)$ represents the $K$ value at epoch $t$, $n$ denotes the total number of nodes in the graph, $p$ is the predefined mask ratio, and $T$ represents the total number of training epochs. This approach enables the model to progressively concentrate on more challenging nodes, thereby enhancing its acquisition of complex structural information.

% This approach gradually provides visible nodes for the model to predict the masked nodes throughout the GMAE training process.
% \text{\large \ding{202}}

\subsection{Overall StructMAE Architecture}
\label{sec:overall-arch}

\paragraph{Training Process.} The StructMAE training process begins with an input graph from which a specified proportion of nodes is chosen based on our selective masking strategy. The selected nodes are subsequently masked using a mask-token. The graph, now with partially masked features, is subsequently fed into the encoder, which generates encoded representations of the nodes. After that, the decoder module is responsible for predicting and reconstructing the features of masked nodes.  For optimization, we adopt the scaled cosine error as utilized in GraphMAE~\cite{graphmae}.

\paragraph{Inference and Downstream Tasks.} StructMAE is designed to cater to two distinct downstream applications: unsupervised and transfer learning.  In unsupervised learning, the encoder processes the input graph without masking during the inference stage.  The node embeddings generated by the encoder are then utilized for graph classification tasks with linear classifiers or support vector machines. In the transfer learning context, the pre-trained models are fine-tuned on different datasets. This fine-tuning enables the model to adjust to new data domains, leveraging the foundational knowledge gained during its initial training on the source dataset. Each of these downstream tasks emphasizes the versatility and applicability of StructMAE in diverse graph learning scenarios.

\begin{table*}[!t]
\centering
\renewcommand\arraystretch{1.35} % 行间距
\setlength\tabcolsep{4pt} % 列间距
\resizebox{1.0\textwidth}{!}{%
\begin{tabular}{@{}l|ccccccc|c@{}}
\toprule
\multirow{1}{*}{} & \multirow{1}{*}{\textbf{PROTEINS}} & \multirow{1}{*}{\textbf{NCI1}} & \multirow{1}{*}{\textbf{IMDB-B}} &  \multirow{1}{*}{\textbf{IMDB-M}} & \multirow{1}{*}{\textbf{COLLAB}} & \multirow{1}{*}{\textbf{REDDIT-B}} & \multirow{1}{*}{\textbf{MUTAG}} & \multirow{1}{*}{\textbf{A.R.}}                                             \\ \midrule
%\midrule
%\rowcolor{Gray}
\multicolumn{8}{c}{\textit{Supervised Methods}}\\
\midrule
% GCN~\cite{gcn}               & $79.68_{\pm 2.05}$    & $71.7_{\pm 4.7}$        & $73.4_{\pm 10.8}$      & $74.3_{\pm 4.6}$           & $71.92_{\pm 1.18}$     & $71.89_{\pm 0.33}$    & $69.50_{\pm 0.98}$  & $0.367_{\pm 0.011}$      \\
GIN~\cite{gin}            & $76.2_{\pm 2.8}$    & $82.7_{\pm 1.7}$              & $75.1_{\pm 5.1}$                 & $52.3_{\pm 2.8}$     & $80.2_{\pm 1.9}$    & $92.4_{\pm 2.5}$ & $89.4_{\pm 5.6}$ & --       \\
DiffPool~\cite{diffpool}                & --       & $92.1_{\pm 2.6}$      & $72.6_{\pm 3.9}$                 & --     & $78.9_{\pm 2.3}$    & $92.1_{\pm 2.6}$      & $75.1_{\pm 3.5}$ & -- \\
\midrule
\multicolumn{8}{c}{\textit{Self-supervised Methods}} \\
\midrule
Infograph~\cite{infoGraph:}               & $74.44_{\pm 0.31}$                                                 & $76.20_{\pm 1.06}$                & $73.03_{\pm 0.87}$                         & $49.69_{\pm 0.53}$                       &$70.65_{\pm 1.13}$                       & $82.50_{\pm 1.42}$  & $\underline{89.01_{\pm 1.13}}$ & $6.86 $                                    \\
GraphCL~\cite{graphCL}               & $74.39_{\pm 0.45}$                                                 & $77.87_{\pm 0.41}$                    & $71.14_{\pm 0.44}$                      & $48.58_{\pm 0.67}$                       &$71.36_{\pm 1.15}$                       & $\underline{89.53_{\pm 0.84}}$ & $86.80_{\pm 1.34}$ & ${{7.43}} $                                        \\
JOAO~\cite{joao}        & $74.55_{\pm 0.41}$                                              & $78.07_{\pm 0.47}$                                  &  $70.21_{\pm 3.08}$              &  $49.20_{\pm 0.77}$                      &     $69.50_{\pm 0.36}$                 &  $85.29_{\pm 1.35}$  & $87.35_{\pm 1.02}$ & ${{8.00}} $                                        \\ 
GCC~\cite{gcc}        & --                                             & --                              &  $72.0$             &  $49.4$                      &    $78.9$                   &  $\textbf{89.8}$  &    -- & ${{5.25}} $                                  \\ 
% MVGRL~\cite{mvgrl}               & --                                                 & --       & $89.70_{\pm 1.10}$              & $74.20_{\pm 0.70}$                       & $51.20_{\pm 0.50}$                       &--                       & $84.50_{\pm 0.60}$                    &$0.094_{\pm 0.008}$                       \\
InfoGCL~\cite{infogcl}               & --                                                 & $80.20_{\pm 0.60}$                    &$75.10_{\pm 0.90}$                        & $51.40_{\pm 0.80}$                       &$80.00_{\pm 1.30}$                       & -- & $\mathbf{91.20}_{\pm \mathbf{1.30}}$    & ${{4.00}} $                        \\
SimGRACE~\cite{simGrace}               & $75.35_{\pm 0.09}$                                                 & $79.12_{\pm 0.44}$                     &$71.30_{\pm 0.77}$                        & --                       &$71.72_{\pm 0.82}$                       & $89.51_{\pm 0.89}$  & $89.01_{\pm 1.31}$ & ${{5.00}} $                                        \\
GraphMAE~\cite{graphmae}        & $75.30_{\pm 0.39}$                                              & $80.40_{\pm 0.30}$                               &  $75.52_{\pm 0.66}$              &  ${51.63}_{\pm 0.52}$                     &   $80.32_{\pm 0.46}$                    &  $88.01_{\pm 0.19}$  &    $88.19_{\pm 1.26}$ & ${{4.43}} $                               \\ 
S2GAE~\cite{s2gae}        & ${\underline{76.37_{\pm {0.43}}}}$                                              & $80.80_{\pm 0.24}$                              &  ${\underline{75.76_{\pm 0.62}}}$              &  $\underline{51.79_{\pm 0.36}}$                     &   $\underline{81.02_{\pm 0.53}}$                    &  $87.83_{\pm 0.27}$  &    $88.26_{\pm 0.76}$  & $\underline{{3.14}} $                               \\ 
\midrule
\midrule
StructMAE-P (ours)      & ${{75.97}}_{\pm 0.38} $   &  ${\textbf{81.91}}_{\pm \mathbf{0.31}}$          & ${{75.72}}_{\pm 0.36}$      &                        ${ {51.25}}_{\pm 0.64}$ &   ${ {80.53}}_{\pm 0.22}$   &  ${{88.25}}_{\pm 0.40}$ &                         ${{87.91}}_{\pm 0.39}$ & $3.71 $                                 \\
StructMAE-L  (ours)    & $\mathbf{76.62}_{\pm \mathbf{0.84}} $   &  ${\underline{81.25_{\pm 1.37}}}$          & ${ \mathbf{75.84}_{\pm \mathbf{0.46}}}$      &                        ${ \textbf{52.05}_{\pm \textbf{0.73}}}$ &   ${ \textbf{81.46}_{\pm \mathbf{0.27}}}$   &  ${89.03_{\pm 0.40}}$ &                         ${88.43_{\pm 0.54}}$ & $\textbf{1.86}$                                \\ \bottomrule
\end{tabular}
}
% \begin{minipage}{1.0\linewidth} \small % \footnotesize %\scriptsize %\scriptsize %\tiny
% \vspace{0.5em}
% Notations:  The results of baselines are from previous paper if avaliable.
% \end{minipage} 
% \vspace{-0.9em}
\caption{Experimental results for \textbf{unsupervised representation learning} in graph classification.  The results for baseline methods are sourced from prior studies. \textbf{Bold} or \underline{underline} indicates the best or second-best result, respectively, among self-supervised methods. \textbf{A.R.} denotes the average rank of self-supervised methods.}
% \vspace{-0.5em}
\label{tab:unspervise}
\end{table*}

% \usepackage{booktabs}
% \usepackage{graphicx}
\begin{table*}[!t]
\centering
\renewcommand\arraystretch{1.30} % 行间距
\setlength\tabcolsep{6pt} % 列间距
\resizebox{\textwidth}{!}{%
\begin{tabular}{@{}l|cccccccc|c@{}}
\toprule
            & \textbf{BBBP}                                  & \textbf{Tox21}                                  & \textbf{ToxCast}                               & \textbf{SIDER}                                 & \textbf{ClinTox}                                  & \textbf{MUV}                          & \textbf{HIV}                                   & \textbf{BACE}                                                       & \textbf{Avg.}               \\ \midrule
No-pretrain & $65.5_{\pm 1.8}$                        & $74.3_ {\pm 0.5}$                         & $63.3_ {\pm 1.5}$                        & $57.2_ {\pm 0.7}$                        & $58.2_ {\pm 2.8}$                           & $71.7_ {\pm 2.3}$               & $75.4_ {\pm 1.5}$                        & \multicolumn{1}{c|}{$70.0_ {\pm 2.5}$}                        & 67.0               \\ \midrule
ContextPred~\cite{pretrain-gnn} & $64.3_ {\pm 2.8}$                        & $75.7_ {\pm 0.7}$             & $63.9_ {\pm 0.6}$                        & $60.9_ {\pm 0.6}$                        & $65.9_ {\pm 3.8}$                           & $75.8_ {\pm 1.7}$               & $77.3_ {\pm 1.0}$                        & \multicolumn{1}{c|}{$79.6_ {\pm 1.2}$}                        & 70.4               \\
AttrMasking~\cite{pretrain-gnn} & $64.3_ {\pm 2.8}$                        & $\underline{{76.7}_ {\pm 0.4}}$  & $64.2_ {\pm \mathbf{0.5}}$ & $61.0_ {\pm 0.7}$            & $71.8_ {\pm 4.1}$                           & $74.7_ {\pm 1.4}$               & $77.2_ {\pm 1.1}$                        & \multicolumn{1}{c|}{$79.3_ {\pm 1.6}$}                        & 71.1               \\
Infomax~\cite{pretrain-gnn}     & $68.8_ {\pm 0.8}$                        & $75.3_ {\pm 0.5}$                         & $62.7_ {\pm 0.4}$                        & $58.4_ {\pm 0.8}$                        & $69.9_ {\pm 3.0}$                           & $75.3_ {\pm 2.5}$               & $76.0_ {\pm 0.7}$                        & \multicolumn{1}{c|}{$75.9_ {\pm 1.6}$}                        & 70.3               \\
GraphCL~\cite{graphCL} & $69.7_ {\pm 0.7}$                        & $73.9_ {\pm 0.7}$                         & $62.4_ {\pm 0.6}$                        & $60.5_ {\pm 0.9}$                        & $76.0_ {\pm 2.7}$                           & $69.8_ {\pm 2.7}$               & $\mathbf{78.5}_ {\pm \mathbf{1.2}}$ & \multicolumn{1}{c|}{$75.4_{ \pm 1.4}$}                        & 70.8               \\
JOAO~\cite{joao}        & $70.2_ {\pm 1.0}$                        & $75.0_ {\pm 0.3}$                         & $62.9_ {\pm 0.5}$                        & $60.0_ {\pm 0.8}$                        & $81.3_ {\pm 2.5}$               & $71.7_ {\pm 1.4}$               & $76.7_ {\pm 1.2}$                        & \multicolumn{1}{c|}{$77.3_ {\pm 0.5}$}                        & 71.9               \\
GraphLoG~\cite{graphlog}    & $\underline{72.5 _{\pm 0.8}}$ & $75.7_ {\pm 0.5}$ & $63.5_ {\pm 0.7}$                        & $61.2_ {\pm 1.1}$ & $76.7_ {\pm 3.3}$             & $76.0_ {\pm 1.1}$   & $77.8_ {\pm 0.8}$            & \multicolumn{1}{c|}{$\underline{{83.5}_ {\pm {1.2}}}$} & $73.4$ \\
RGCL~\cite{rgcl}    & $71.2 _{\pm 0.9}$ & $75.3_ {\pm 0.5}$ & $63.1_ {\pm 0.3}$                        & $61.2_ {\pm 0.6}$ & $85.0_ {\pm 0.8}$             & $73.1_ {\pm 1.2}$   & $77.3_ {\pm 0.8}$            & \multicolumn{1}{c|}{${75.7}_ {\pm {1.3}}$} & $72.7$ \\
GraphMAE~\cite{graphmae}    & $72.0_ {\pm 0.6}$            & $75.5_ {\pm 0.6}$             & $64.1_ {\pm 0.3}$            & $60.3_ {\pm 1.1}$                        & $82.3_ {\pm 1.2}$ & $76.3_ {\pm 2.4}$ & $77.2_ {\pm 1.0}$                        & $83.1_ {\pm 0.9}$                                 & $73.8$ \\  
GraphMAE2~\cite{graphmae2}    & $71.6_ {\pm 1.6}$            & $75.9_ {\pm 0.8}$             & $\mathbf{65.6_ {\pm 0.7}}$           & $59.6_ {\pm 0.6}$                        & $78.8_ {\pm 3.0}$ & $\underline{78.5_ {\pm 1.1}}$ & $76.1_ {\pm 2.2}$                        & $81.0_ {\pm 1.4}$ & $73.4$\\     
% S2GAE~\cite{s2gae}    & $67.6_ {\pm 2.0}$            & $69.6_ {\pm 1.0}$             & $58.7_ {\pm 0.8}$            & $55.4_ {\pm 1.3}$                        & $59.6_ {\pm 1.1}$ & $60.1_ {\pm 2.4}$ & $68.0_ {\pm 3.7}$                        & $68.6_ {\pm 2.1}$  
% & $63.5$ \\  
Mole-BERT~\cite{molebert}    & $71.9_ {\pm 1.6}$            & $\mathbf{76.8_ {\pm 0.5}}$             & $64.3_ {\pm 0.2}$            & $\mathbf{62.8_ {\pm 1.1}}$                       & $78.9_ {\pm 3.0}$ & $\mathbf{78.6_ {\pm 1.8}}$ & $78.2_ {\pm 0.8}$                        & $80.8_ {\pm 1.4}$                                 & $74.0$ \\\midrule \midrule
StructMAE-P (ours)   & $\mathbf{72.6_ {\pm 0.9}}$           & $75.8_ {\pm 0.4}$             & $\underline{64.5_ {\pm 0.5}}$            & $\underline{62.0_ {\pm 0.4}}$                        & $\underline{86.0_ {\pm 1.6}}$ & $77.7_ {\pm 1.1}$ & $77.4_ {\pm 1.0}$                        & $\mathbf{84.3_ {\pm \mathbf{0.6}}}$                                  & $\underline{75.0}$ \\
StructMAE-L (ours)   & $\underline{72.5_ {\pm 0.9}}$            & $75.3_ {\pm 0.4}$             & $64.0_ {\pm 0.4}$            & $61.3_ {\pm 0.5}$                        & $\mathbf{87.9_ {\pm \mathbf{2.1}}}$ & $78.0_ {\pm 1.1}$ & $\underline{78.3_ {\pm 0.8}}$                        & $83.2_ {\pm 0.9}$                                 & $\mathbf{75.1}$ \\
\bottomrule
\end{tabular}%
}
\caption{Experimental results for \textbf{transfer learning} on molecular property prediction. The model is initially pre-trained on the ZINC15 dataset and subsequently fine-tuned on the above datasets. The reported metrics are ROC-AUC scores. The results for baseline methods are derived from prior studies. \textbf{Bold} or \underline{underline} indicates the best or second-best result, respectively. \textbf{Avg.} denotes the average performance. }
\label{tab:transfer}
\end{table*}

\section{Experiment}
\label{sec:experiment}

% In this section, we conduct extensive experiments  on a variety of benchmark graph datasets to validate the effectiveness of the proposed method. 

\subsection{Unsupervised Representation Learning}

\paragraph{Objective.} To assess the efficacy of the pre-trained model in its feature extraction capability, we subject it to a series of unsupervised tasks. Achieving success in these tasks will highlight the model's proficiency in learning high-quality and informative representations, which are crucial for various downstream graph analytics tasks.

\paragraph{Settings.} \textbf{Datasets.} We employ seven real-world datasets, including MUTAG, IMDB-B, IMDB-M, PROTEINS, COLLAB, REDDIT-B, and NCI1, involving diverse domains and sizes. \textbf{Baseline Models.} To demonstrate the effectiveness of our proposed method, we compare StructMAE with the following 10 baseline models: \textit{1) Two supervised models:} GIN~\shortcite{gin} and DiffPool~\shortcite{diffpool}; \textit{2) Six contrastive models:} Infograph~\shortcite{infoGraph:}, GraphCL~\shortcite{graphCL}, JOAO~\shortcite{joao}, GCC~\shortcite{gcc}, InfoGCL~\shortcite{infogcl}, and SimGRACE~\cite{simGrace}; \textit{3) Two  generative models:} GraphMAE~\cite{graphmae} and S2GAE~\cite{s2gae}. We report the results from previous papers according to graph classification research norms. \textbf{Implementation Details.} In the evaluation protocol, we initially generate graph embeddings using the encoder and readout function. Then, the encoded graph-level representations are fed into a downstream LIBSVM~\cite{svm} classifier for label prediction, consistent with other baseline models. The performance is assessed by measuring the mean accuracy obtained from a 10-fold cross-validation, and this evaluation is repeated five times to ensure robustness.  
%The introduction of datasets, baseline models, and detailed hyper-parameters settings are provided in the Appendix.

\begin{table*}[t]

\centering
\subfloat[
\textbf{Pre-defined metric.} Close. and Between. denote closeness centrality and betweenness centrality, respectively.
\label{tab:abl-predefine}
]{
\begin{minipage}{0.23\linewidth}{\begin{center}
\tablestyle{4pt}{1.05}
\begin{tabular}{y{33}x{23}x{19}}
Metric & IM-M & COL \\
\shline
PageRank &  \baseline{\textbf{51.25}} & \baseline{\textbf{80.53}} \\
Degree & 50.87 & 80.28 \\
Close. & 51.04 & 80.44\\
Between. & 50.97 & 80.17 \\
\end{tabular}
\end{center}}\end{minipage}
}
\hspace{1em}
\subfloat[
\textbf{Scoring function.} M\&G mixes the scores generated by MLP and GNN.
\label{tab:abl-scoring-funciton}
]{
\centering
\begin{minipage}{0.23\linewidth}{\begin{center}
\tablestyle{4pt}{1.05}
\begin{tabular}{y{40}x{20}x{21}}
Function & IM-B & RE-B \\
\shline
MLP  & 75.32 & 88.70 \\
GNN  & 75.52 & 88.83 \\
M\&G & \baseline{\textbf{75.84}} & \baseline{\textbf{89.03}} \\
\end{tabular}
\end{center}}\end{minipage}
}
\hspace{1em}
\subfloat[
\textbf{Masking strategy.} E-to-H is short for easy-to-hard.
\label{tab:abl-masking-strategy}
]{
\begin{minipage}{0.19\linewidth}{\begin{center}
\tablestyle{4pt}{1.05}
\begin{tabular}{y{36}z{18}z{18}}
Strategy & \multicolumn{1}{c}{IM-B} & \multicolumn{1}{c}{RE-B} \\
\shline
Top  & 75.62 & 87.81 \\
Middle  & 75.08 & 75.03 \\
Bottom & 75.28 & 87.82 \\
E-to-H & \baseline{\textbf{75.72}} & \baseline{\textbf{88.35}} \\

\end{tabular}
\end{center}}\end{minipage}
}
\hspace{1em}
\subfloat[
\textbf{Randomness.} Performance of StructMAE with/without random masking.
\label{tab:abl-random-noise}
]{
\begin{minipage}{0.22\linewidth}{\begin{center}
\tablestyle{4pt}{1.05}
\begin{tabular}{x{24}x{20}x{18}}
Dataset & w/o & w \\
\shline
IM-B & 74.92 & \baseline{\textbf{75.72}} \\
IM-M & 50.36 & \baseline{\textbf{51.25}} \\
COL & 79.27 & \baseline{\textbf{80.53}} \\
RE-B & 81.71 & \baseline{\textbf{88.25}} \\
\end{tabular}
\end{center}}\end{minipage}
}
\vspace{-.3em}
\caption{\textbf{Ablation Study}. The best results are in \textbf{bold}. Default settings are marked in \colorbox{baselinecolor}{gray}. IM-B, IM-M, RE-B and COL correspond to IMDB-B, IMDB-M, REDDIT-B and COLLAB, respectively.}
\label{tab:ablations}

\vspace{-7pt}
\end{table*}

\paragraph{Results.} The results are detailed in Table~\ref{tab:unspervise}, where StructMAE-P denotes StructMAE with predefined SBS, and StructMAE-L is with learnable SBS. Analyzing these results enables deriving several observations:  \textbf{1) State-of-the-art Performance:}  StructMAE-L outperforms existing self-supervised baselines on four out of seven datasets. Furthermore, it attains state-of-the-art performance considering the average rank across these datasets. Meanwhile, StructMAE-P  maintains competitive performance against other self-supervised methods. These results emphatically demonstrates the efficacy of the proposed StructMAE approach.  Please note that,  StructMAE focuses solely on masking nodes, whereas S2GAE extends its masking strategy to include edges. The differing approach of S2GAE underscores a potential avenue for further development in StructMAE. \textbf{2) Comparison with Supervised Methods:} Remarkably, the StructMAE-L, though self-supervised, attain comparable or superior performance on certain datasets, including PROTEINS, IMDB-B, and COLLAB. This finding indicates that the representations learned by StructMAE are high-quality and informative, aligning with supervised learning benchmarks. \textbf{3) Comparison with GraphMAE:} In comparison to GraphMAE, which employs a random node masking strategy, StructMAE consistently outperforms it, providing further evidence to support our hypothesis that incorporating structural knowledge into the masking process can significantly enhance the model's learning capabilities. \textbf{4) Predefined vs. Learnable SBS:} A notable trend is that the StructMAE-L's performance generally surpasses that of StructMAE-P, particularly on complex datasets such as COLLAB and REDDIT-B. This trend indicates the greater adaptability and effectiveness of the learnable SBS in handling complex and variable graph structures, as opposed to the predefined method. These results collectively validate the core principles behind StructMAE and its components, highlighting its potential as a powerful tool for unsupervised representation learning in graph-structured data.

\subsection{Transfer Learning}

\paragraph{Objective.} The primary goal of the transfer learning task is to evaluate the transferability of the pre-training scheme utilized in StructMAE. This involves pre-training the model on a specific dataset and fine-tuning it with different datasets.

\paragraph{Settings.} \textbf{Datasets.} During the initial pre-training phase, StructMAE is trained on a dataset comprising two million unlabeled molecules obtained from the ZINC15~\cite{zinc15} dataset. Subsequently, the model is fine-tuned on eight classification benchmark datasets featured in the MoleculeNet dataset~\cite{moleculenet}. In our evaluation, we adopt a scaffold-split approach for splitting the datasets, as outlined in~\cite{graphmae}. \textbf{Baseline Models.} To demonstrate the effectiveness of our proposed method, we compare StructMAE with the following 10 baseline models: \textit{1) Three unsupervised models:} Infomax, AttrMasking and ContextPred~\cite{pretrain-gnn}; \textit{2) Four contrastive models:} GraphCL~\shortcite{graphCL}, JOAO~\shortcite{joao},   GraphLOG~\cite{graphlog}, and RGCL~\cite{rgcl}; \textit{3) Three generative models:} GraphMAE~\cite{graphmae}, GraphMAE2~\cite{graphmae2}, and Mole-BERT~\cite{molebert}. We report the results of baseline models from previous papers according to research norms. \textbf{Implementation Details.} Experiments are conducted 10 times, and the mean and standard deviation of the ROC-AUC scores are reported. According to the default settings used in prior research~\cite{graphmae}, a 5-layer GIN model~\cite{gin} is employed as the encoder and a single-layer GIN as the decoder in our StructMAE framework. 
% The introduction of datasets, baseline models, and detailed hyper-parameters are summarized in the Appendix.

% This configuration complies with standard practices in the field and guarantees a fair comparison with baseline models.

\paragraph{Results.} The detailed results in Table~\ref{tab:transfer} offer insightful observations into StructMAE's performance within the transfer learning context.
\textbf{1) State-of-the-Art Performance Across Datasets: } StructMAE-P and StructMAE-L demonstrate state-of-the-art performance across an ensemble of eight datasets. Specifically, StructMAE-P and StructMAE-L achieved 1.4\% and 1.5\% improvements in average performance metrics, respectively. Additionally, each method individually achieves top-tier performance on several datasets. These results validate the StructMAE's ability to effectively generalize learned representations across diverse datasets.
\textbf{2) Superiority of StructMAE-L:} The results reveal the superior performance of StructMAE-L over StructMAE-P on the average metrics. This trend further highlights the enhanced adaptability and effectiveness of the learnable SBS method, similar to observations made previously in unsupervised learning.

% In summary, StructMAE demonstrates notable success across a range of benchmarks, evidencing its competitive performance in both unsupervised and transfer learning tasks.

\subsection{Ablation Study}
\label{sec:further-discussion}

% \paragraph{Impact of Mask Ratio, curriculum parameter, and the embedding dimension.} We first analyze the effects of $\lambda$ and $n$ using AGP-G on three graph datasets (Citeseer, Pubmed, and Texas) and presented the results in Figure~\ref{fig:analyse}. Specifically, we investigated how the number of layers impacts node classification performance. Our findings indicate that as $l$ increases from low to high values, test accuracy decreases due to over-fitting. Additionally, when $\beta$ is relatively small, our model's accuracy curve remains smooth indicating less sensitivity to hyper-parameters.

% \paragraph{The impact of Pre-define Metrics, Learnable and Scoring Functions, and Random Noise.}  

A detailed study is conducted to evaluate the impact of different components within StructMAE. It is important to note that, except for the components under analysis, all other aspects of the model remain consistent with the comprehensive StructMAE. The findings, as outlined below, provide valuable insights into the significance of each component: \textbf{1) Efficacy of Pre-defined Metrics:} As shown in Table~\ref{tab:abl-predefine}, the PageRank metric consistently demonstrates superior performance compared to other pre-defined metrics. However, other metrics also demonstrate commendable performances, suggesting their potential applications in specific scenarios. \textbf{2)  Scoring Function Comparison: }  As displayed in Table~\ref{tab:abl-scoring-funciton}, we observe that the performance of GNN scoring function outperforms the Multilayer Perceptron (MLP). This outcome emphasizes the importance of incorporating structural information into the scoring process. Furthermore, the combined use of GNN and MLP consistently yields superior performance compared to using either one independently. \textbf{3) Masking Strategy Efficiency: } The results are detailed in Table~\ref{tab:abl-masking-strategy}, where Top, Middle, and Bottom denote the masking of the top, middle, and bottom $p*n$ nodes, respectively, based on the masking probability. Analysis of the results indicates that the easy-to-hard masking strategy consistently surpasses the performance of other methods. This finding supports our perspective on the effectiveness of gradually increasing the learning challenge, validating the strategic design of our training approach. \textbf{4) Role of Random Noise:} Notably, a significant drop in performance is observed in the absence of random noise, as indicated in Table~\ref{tab:abl-random-noise}. This indicates that the inclusion of randomness in the model enables access to a broader range of information and simultaneously strengthens the learning process robustness. 

% Owing to space limitations, the detailed parameter analysis, including the mask ratio, extra probability, embedding hidden dimension, and the number of warm-up epochs, is provided in the Appendix. 

% This analysis comprehensively covers various parameters, including the mask ratio, decay ratio, embedding hidden dimension, and the number of warm-up epochs. 

\section{Conclusion} 
This study proposes the StructMAE model, a novel structure-guided masking strategy that incorporates prior structural knowledge into the masking process, thereby enhancing the pre-training model's learning efficiency. The StructMAE framework consists of two pivotal steps: SBS and SGM. Extensive experiments, encompassing two distinct graph learning tasks, demonstrate that StructMAE significantly outperforms existing self-supervised pre-training methods. These results highlight our approach's effectiveness in leveraging structural information for improved model performance. Despite its competitive performance, StructMAE still has room for improvement. For instance, 1) devising more effective scoring methods to fully exploit the structural information, 2) extending the structure-guided masking strategy to encompass edge masking, and 3) expanding structure-guided masking to a broader spectrum of tasks (\textit{e.g.}, node classification).

\section*{Acknowledgments}
The work of Wenbin Hu was supported by the National Key Research and Development Program of China (2023YFC2705700). This work was supported in part by the Natural Science Foundation of China (No. 82174230), Artificial Intelligence Innovation Project of Wuhan Science and Technology Bureau (No. 2022010702040070), Natural Science Foundation of Shenzhen City (No. JCYJ20230807090211021).

\bibliographystyle{named_short}
\bibliography{reference_short}

\clearpage
\appendix

\section{Parameter Analysis}
\begin{figure}[!h] % !htb
\setlength{\abovecaptionskip}{-0.0cm}   %调整图片标题与图距离
\setlength{\belowcaptionskip}{-0.2cm}   %调整图片标题与下文距离
\begin{center}
\includegraphics[width=0.99\linewidth]{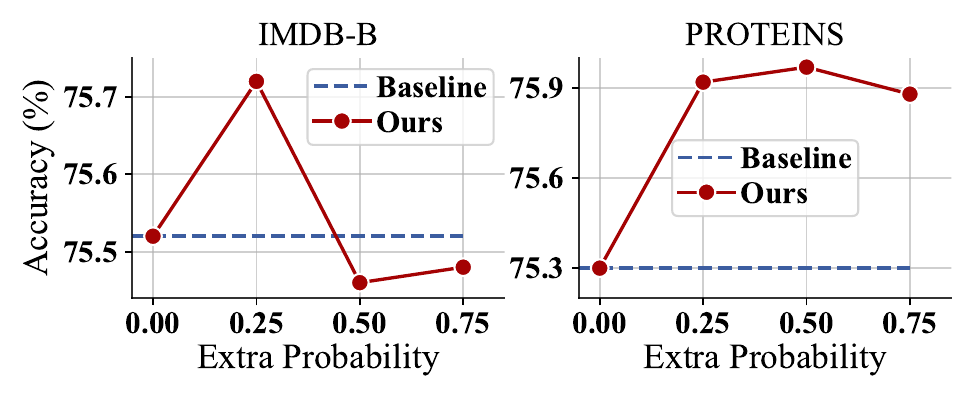}
\end{center}
\caption{Performance of \textbf{StructMAE-P} (Ours) with different extra probabilities. Baseline refers to  GraphMAE~\protect\cite{graphmae}.}
\label{fig:analysis-extra}
\end{figure}

In this section, we investigate the effects of varying the extra masking probability assigned to nodes deemed structurally significant in the graph. The detailed results are presented in Figure~\ref{fig:analysis-extra}. Analyzing the results, we aim to elucidate how different levels of emphasis on structural information during the masking process affect the model's training performance. Our findings reveal a notable trend: the model typically achieves peak accuracy when the additional masking probability is set to either $0.25$ or $0.50$. This observation implies the existence of an optimal range for integrating structural information into the masking process. When properly balanced, this integration enhances the model's training performance, enabling it to focus more effectively on key nodes within the graph. However, deviating beyond this optimal range can lead to detrimental effects. Excessive emphasis on structurally significant nodes, indicated by higher extra masking probabilities, appears to impede the model's learning process. This could result from the model becoming overly biased towards certain nodes, potentially overlooking other valuable information distributed across the graph.

\section{Visualization} 

\begin{figure}[!h] \centering
    \includegraphics[width=0.49\textwidth]{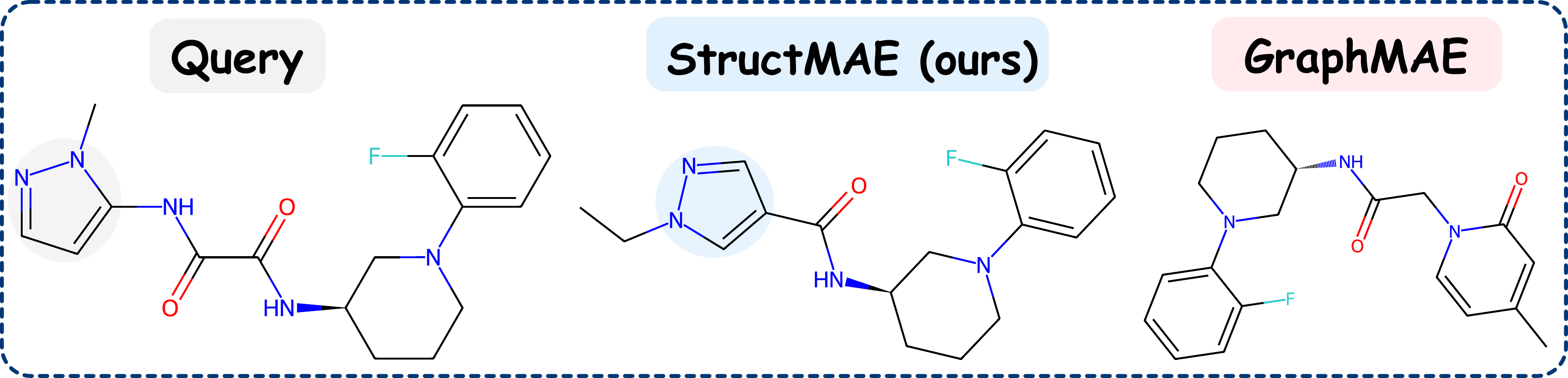}
    \caption{Visualization of the top-ranked (Top@1) molecule, identified by molecular representation similarity, to the query molecule from ZINC15. The molecule representations are obtained from the pre-trained model in the transfer learning task.} \label{fig:img1}
\end{figure}

For a deeper exploration of the structure information embedded in the pre-trained representations generated by StructMAE (trained on the ZINC15 dataset), we assess the similarity (\textit{i.e.}, cosine similarity) between a given query molecule and other molecules, displaying the top-most similar molecules. As shown in Figure~\ref{fig:img1}, the molecule ranked highest by StructMAE (Top@1) demonstrates a significant similarity to the query molecule, mirroring both the types of nodes and the graph structure. In contrast, the Top@1 molecule identified by GraphMAE~\cite{graphmae} shares only atom types with the query molecule and is significantly deficient in replicating essential structural features (\textit{e.g.}, the subgraphs highlighted in \textcolor{gray}{grey} or  \textcolor{blue}{blue} colors). Therefore, this illustrates that our structure-guided masking can assist the model in more accurately capturing the graph structural information.

\begin{table*}[!t]
    \setlength\tabcolsep{6pt} % default value: 6pt
    \centering
    \footnotesize
    \renewcommand\arraystretch{1.4} % 行间距
    \setlength\tabcolsep{5pt} % 列间距
    \begin{tabular}{c|c|ccccccc}\toprule
    &Dataset&\textbf{IMDB-B} &\textbf{IMDB-M} &\textbf{PROTEINS} &\textbf{COLLAB} &\textbf{MUTAG}  &\textbf{REDDIT-B} &\textbf{NCI1}
    % &\textbf{Predict level} &\textbf{Predict task} &\textbf{Metric} 
    \\\midrule
    \multirow{12}{*}{\parbox{1.5cm}{\centering {Hyper-\\parameters}}}
    &Mask ratio &0.5 &0.25 &0.25 &0.75 &0.9 &0.75 &0.3
    % &graph &2-class classif. & Accuracy 
    \\
    &Hidden\_size &512 &512 &512 &256 &32 &512 &512 
    \\
    &Encoder&GIN&GIN&GIN&GIN&GIN&GCN&GIN\\
    &Decoder&GIN&GIN&GIN&GIN&GIN&GCN&GIN\\
    &Num\_layers &2&2&3&2&5&2&3\\
    &Learning rate &0.00015&0.005&0.00015&0.00015&0.0005&0.005&0.005\\
    &Weight\_decay &0 &0 &0 &0&0 &0 &0\\
    &Batch\_size &32&32 &32 &32 &64 &8 &16\\
    % &Max\_epoch &100&50 &150 &20 &25 &150 &300\\
    &Pooling &mean&mean&max&max&sum&max&sum\\
    
    &Extra Probability $\beta$ &0.25 &0.85 &0.5 &0.5 &0.85 &0.25 &0.25 \\
    \bottomrule
    \end{tabular}
     \caption{Hyper-parameters in \textbf{StructMAE-P} in \textbf{unsupervised }representation learning.}
    \label{tab:hyper-p}
\end{table*}

\begin{table*}[!t]
    \setlength\tabcolsep{6pt} % default value: 6pt
    \centering
    \footnotesize
    \renewcommand\arraystretch{1.4} % 行间距
    \setlength\tabcolsep{5pt} % 列间距
    \begin{tabular}{c|c|ccccccc}\toprule
    &Dataset&\textbf{IMDB-B} &\textbf{IMDB-M} &\textbf{PROTEINS} &\textbf{COLLAB} &\textbf{MUTAG}  &\textbf{REDDIT-B} &\textbf{NCI1}
    % &\textbf{Predict level} &\textbf{Predict task} &\textbf{Metric} 
    \\\midrule
    \multirow{13}{*}{\parbox{1.5cm}{\centering {Hyper-\\parameters}}}
&Mask ratio &0.5 &0.25 &0.25 &0.5 &0.3 &0.75 &0.25
    % &graph &2-class classif. & Accuracy 
    \\
    &Hidden\_size &512 &512 &512 &256 &32 &512 &512 
    \\
    &Encoder&GIN&GIN&GIN&GIN&GIN&GCN&GIN\\
    &Decoder&GIN&GIN&GIN&GIN&GIN&GCN&GIN\\
    &Num\_layers &2&3&3&2&5&2&3\\
    &Learning rate &0.00015&0.005&0.00015&0.00015&0.0005&0.005&0.005\\
    &Weight\_decay &0 &0 &0 &0&0 &0 &0\\
    &Batch\_size &32&32 &32 &32 &64 &8 &16\\
    % &Max\_epoch &100&70 &100 &25 &20 &125 &350\\
    &Pooling &mean&mean&max&max&sum&max&sum\\
    
    &Extra Probability $\beta$ &0.75 &0.75 &0.85 &0.5 &0.25 &0.25 &0.85 \\
    &Balance Parameter $\alpha$ &0.1&0.1&100.0&100.0&10.0&0.1&10.0\\
    % \\\midrule
    % \multirow{3}{*}{\parbox{3cm}{\centering\textbf{Pretrain-GNNs}\\\cite{pretrain-gnn}}}
    % &ZINC & 2,000,000 &23.2 &24.9 &graph &regression &MAE \\
    % &PATTERN &14,000 &118.9 &3,039.3 &inductive node &2-class classif. &Accuracy \\
    % &CLUSTER &12,000 &117.2 &2,150.9 &inductive node &6-class classif. &Accuracy\\
    \bottomrule
    \end{tabular}
     \caption{Hyper-parameters in \textbf{StructMAE-L} in \textbf{unsupervised }representation learning.}
    \label{tab:hyper-l}
\end{table*}

% \paragraph{Synthetic Datasets.} In the main text, we conduct experiments on synthetic datasets to illustrate the superiority of Gapformer in complexity. In this section, we introduce in detail how to generate synthetic graphs. Specifically, we generate synthetic graphs in various node sizes by adopting the \textbf{FakeDataset} approach from PyG\footnote{\url{https://pytorch-geometric.readthedocs.io/en/latest/_modules/torch_geometric/datasets/fake.html\#FakeDataset}}. We set the average degree to 10, and the feature channels to 64. Then, we generate graphs with nodes ranging from 100 to 2,500. We present the code of generating synthetic datasets in Figure~\ref{fig:code}. 

% \begin{figure}[!h] % !htb
% \setlength{\abovecaptionskip}{-0.1cm}   %调整图片标题与图距离
% \setlength{\belowcaptionskip}{-0.3cm}   %调整图片标题与下文距离
% \begin{center}
% \includegraphics[width=0.92\linewidth]{pics/code.pdf}
% \end{center}
% \caption{PyTorch implementation of generating synthetic datasets.}
% \label{fig:code}
% \end{figure}

\section{Implementation Details}
\label{sec:implementaion-detail}

% We evaluate the effectiveness of the proposed model in terms of graph classification accuracy. We conduct 10 trials with random seeds for each model. For each baseline, we refer to the recommended settings in the official implementations.  We implement Gapformer with Python (3.7.0), Pytorch (1.11.0), and Pytorch Geometric (2.2.0). All experiments are conducted on a Linux server with two NVIDIA A100s.

\subsection{Details for Experiments on Unsupervised Representation Learning}
\paragraph{Environment.}
In \textbf{unsupervised} representation learning, we implement StructMAE with Python (3.8), Pytorch (2.0.0), scikit-learn (1.3.2) and Pytorch Geometric (2.4.0). All experiments are conducted on a Linux server with 8 NVIDIA A100 GPUs. 

\paragraph{Training Details.} The Adam~\cite{adam} optimizer and batch normalization are employed for both StructMAE-P and StructMAE-L. The mask ratio is searched within $\{0.25, 0.5, 0.75\}$ for most datasets, and the additional masking probability (denoted as $\beta$ in the main manuscript) is searched within $\{0.25, 0.5, 0.75, 0.85\}$. The hidden dimension is set as 512 for most datasets, except 32 for MUTAG and 256 for COLLAB. Particularly for StructMAE-L, the ratio of warm-up epochs is set as 0.0 for most datasets, except 0.2 for PROTEINS. Besides, the balance parameter between GNN and MLP is searched within $\{0.0, 0.1, 1.0, 10.0, 100.0\}$. In terms of the GNN function used in the scoring process, a one-layer GIN~\cite{gin} is employed for most datasets. However, for MUTAG, we use a GCN~\cite{gcn} to better suit its specific structure and properties. For a comprehensive understanding of our model configurations, detailed hyperparameters for StructMAE-P and StructMAE-L are presented in Table \ref{tab:hyper-p} and Table \ref{tab:hyper-l}, respectively. 

\begin{table*}[!t]
    \setlength\tabcolsep{6pt} % default value: 6pt
    \centering
    \footnotesize
    \renewcommand\arraystretch{1.4} % 行间距
    \setlength\tabcolsep{5pt} % 列间距
    \begin{tabular}{c|c|cccccccc}\toprule
    &&\textbf{BBBP} &\textbf{Tox21} &\textbf{ToxCast} &\textbf{SIDER} &\textbf{ClinTox}  &\textbf{MUV} &\textbf{HIV} &\textbf{BACE} 
    % &\textbf{Predict level} &\textbf{Predict task} &\textbf{Metric} 
    \\\midrule
    \multirow{3}{*}{\parbox{2.0cm}{\centering{\textbf{StructMAE-P}}}}
    &Batch size &128 &128 &32 &128 &32 &128 &128 &32
    \\
    &Learning rate &0.005 &0.001 &0.005 &0.001 &0.001 &0.001 &0.001 &0.001
    \\
    &Dropout ratio &0.6 &0.5 &0.5 &0.5 &0.4 &0.5 &0.5 &0.5
    \\
    % &Max\_epoch &60 &140 &40 &20 &150 &20 &120 &100\\
    % &Earlystop &False&True&True&False&False&True&False&False
    %  \\
     \midrule
    \multirow{3}{*}{\parbox{2.0cm}{\centering {\textbf{StructMAE-L}}}}
        &Batch size &64 &32 &32 &128 &32 &128 &128 &32
    \\
    &Learning rate &0.001 &0.005 &0.005 &0.001 &0.001 &0.001 &0.001 &0.001
    \\
    &Dropout ratio &0.6 &0.6 &0.5 &0.5 &0.5 &0.5 &0.6 &0.4\\
    % &Max\_epoch &120 &40 &100 &20 &110 &20 &170 &40\\
    % &Earlystop &False&False&True&False&False&True&False&False \\
    % \\\midrule
    % \multirow{3}{*}{\parbox{3cm}{\centering\textbf{Pretrain-GNNs}\\\cite{pretrain-gnn}}}
    % &ZINC & 2,000,000 &23.2 &24.9 &graph &regression &MAE \\
    % &PATTERN &14,000 &118.9 &3,039.3 &inductive node &2-class classif. &Accuracy \\
    % &CLUSTER &12,000 &117.2 &2,150.9 &inductive node &6-class classif. &Accuracy\\
    \bottomrule
    \end{tabular}
     \caption{Hyper-parameters in the finetuning phase of \textbf{transfer} learning.}
    \label{tab:transfer-finetune}
\end{table*}

% \usepackage{multirow}
% \usepackage{graphicx}
\begin{table*}[!h]
    \setlength\tabcolsep{6pt} % default value: 6pt
    \centering
    \footnotesize
    \renewcommand\arraystretch{1.4} % 行间距
    \setlength\tabcolsep{10pt} % 列间距
    \begin{tabular}{c|cccccc}\toprule
    \textbf{Dataset} &\textbf{\# Graphs} &\textbf{Avg. \# nodes} &\textbf{Avg. \# edges} &\textbf{Predict level} &\textbf{Predict task} &\textbf{Metric} 
    % &\textbf{Predict level} &\textbf{Predict task} &\textbf{Metric} 
    \\\midrule
    % \multirow{7}{*}
    NCI1 & 4,110 & 29.8 & 32.3 &graph &2-class classif.& Accuracy 
    % &graph &2-class classif. & Accuracy 
    \\
    PROTEINS & 1,113 & 39.1 & 72.8 &graph &2-class classif.& Accuracy 
    % &graph &2-class classif. & Accuracy 
    \\
    MUTAG & 188 & 17.9  & 19.7  &graph &2-class classif.& Accuracy 
    % &graph &2-class classif. & Accuracy 
    \\
    COLLAB & 5,000 & 74.5  & 2,457.7 &graph &3-class classif.& Accuracy 
    % &graph &3-class classif. & Accuracy 
    \\
    IMDB-B & 1,000 & 19.8  & 96.5 &graph &2-class classif.& Accuracy 
    % &graph &2-class classif. & Accuracy 
    \\
    IMDB-M & 1,500 & 13.0  & 65.9 &graph &3-class classif.& Accuracy 
    \\
    REDDIT-B & 2,000 & 429.7  & 497.8 &graph &2-class classif.& Accuracy 
    \\
    % \\\midrule
    % \multirow{3}{*}{\parbox{3cm}{\centering\textbf{Pretrain-GNNs}\\\cite{pretrain-gnn}}}
    % &ZINC & 2,000,000 &23.2 &24.9 &graph &regression &MAE \\
    % &PATTERN &14,000 &118.9 &3,039.3 &inductive node &2-class classif. &Accuracy \\
    % &CLUSTER &12,000 &117.2 &2,150.9 &inductive node &6-class classif. &Accuracy\\
    \bottomrule
    \end{tabular}
     \caption{Overview of the datasets used in \textbf{unsupervised} representation learning.}
    \label{tab:unsupervised-datasets}
\end{table*}

\paragraph{Evaluation.}
During the evaluation phase, we focus on the generation and utilization of graph embeddings for the classification task. \textbf{\text{ \Large \ding{202}} Generation of Graph Embeddings:} We employ the encoder and readout function of our model to create graph embeddings. 
\textbf{\text{ \Large \ding{203}} Classification using LIBSVM:} The generated embeddings are subsequently fed into a LIBSVM \cite{svm} classifier. This approach is in line with the practices adopted by other baseline models ~\cite{graphmae,s2gae} in the field. \textbf{\text{ \Large \ding{204}} Performance Assessment:} To assess the performance of our model, we utilize mean accuracy as the primary metric. This accuracy is derived from a 10-fold cross-validation process, ensuring a comprehensive evaluation. To enhance the robustness and reliability of our results, this 10-fold cross-validation is repeated five times.
\textbf{\text{ \Large \ding{205}} Baseline Configuration:} Each of the baseline models included in our comparison is configured based on the recommended settings provided in their official implementations.

% We fix some hyperparameters for the convenience of the tuning work: the dropout is set to 0.5, the weight decay is set to $5e^{-4}$, and the dimension of position encodings is set to 20.  The hidden dimension is searched within $\{64, 128, 256 \}$. The other hyperparameters are set as follows (we use the same notation as the original papers).

% \begin{itemize}
%   \item For UniMP~\cite{unimp}, the learning rate is set to $0.01$, and the label rate is set to $0.65$.
%   \item For ANS-GT\cite{ans-gt}, the weight decay is set to $0.01 $, and the batch size is set to $32$.
%   \item For NAGphormer~\cite{nagphormer}, the weight decay is set to $5e^{-5}$, the batch size is set to $2000$, the hop $k$ is set to 5, and the readout is attention-sum.
%   \item For our Gapformer, the pooling ratio is set as $0.1$,  the balance parameter is searched within $\{0.1, 1 \}$, and the hop $k$ is searched within $\{2, 3\}$.
% \end{itemize}

\subsection{Details for Experiments on Transfer Learning} 
\paragraph{Environment.}
In \textbf{transfer} learning, we implement StructMAE with Python (3.8), Pytorch (1.13.1), scikit-learn (1.3.2), rdkit (2022.03.2) and Pytorch Geometric (2.0.3). All experiments are conducted on a Linux server with 8 NVIDIA A100 GPUs. 

\paragraph{Training Details.}

In the transfer learning task, the experimental setup for StructMAE closely follows the configurations used in GraphMAE~\cite{graphmae}. We have made specific adjustments to the mask ratio and introduced new hyperparameters to tailor the pre-training and finetuning phases to our research objectives. \textbf{1) Pre-Training Phase:}  For StructMAE-P, we set the mask ratio at $0.5$. Additionally, the extra probability, denoted as $\beta$, is configured to $0.25$.   In the case of StructMAE-L (Learnable Version), the mask ratio and $\beta$ are both set to $0.5$. Furthermore, the balance parameter, represented as $\alpha$, is fixed at $1.0$. \textbf{2) Finetuning Phase:}  During the finetuning phase, we employ the Adam optimizer to refine the model's performance further. Key hyperparameters, including the learning rate, batch size, and dropout ratio, are tuned to optimize the model for each specific dataset. 
The learning rate is varied among $\{0.001, 0.005\}$; the batch size are selected from $\{32, 64, 128\}$;  the dropout ratio is adjusted between $\{0.4, 0.5, 0.6\}$. To provide a comprehensive view of our finetuning phase, detailed hyperparameter configurations are presented in Table \ref{tab:transfer-finetune}.

\paragraph{Evaluation.} In the finetuning phase, we adopt a scaffold-split approach for splitting the datasets and report the mean and standard deviation of ROC-AUC scores in 10 experiments following the guidelines presented in~\cite{graphmae}.

\begin{table}[!t]
\centering
\renewcommand\arraystretch{1.4} % 行间距
\setlength\tabcolsep{2.5pt} % 列间距
\resizebox{0.48\textwidth}{!}{%
\begin{tabular}{@{}c|c@{}}
\toprule
\textbf{ Models }                & \textbf{ Code Links}                                                                                                 \\ \midrule
GIN~\shortcite{gin} & \url{https://github.com/weihua916/powerful-gnns}                               \\
Diffpool~\shortcite{diffpool}                    & \url{https://github.com/RexYing/diffpool}                                                 \\
% Graphormer~\shortcite{graphormer-v1}             & \url{https://github.com/Microsoft/Graphormer}                                             \\  
\midrule
Infograph~\shortcite{infoGraph:}                 & \url{https://github.com/sunfanyunn/InfoGraph}                                                \\
GraphCL~\shortcite{graphCL}                  & \url{https://github.com/Shen-Lab/GraphCL} \\
JOAO~\shortcite{joao}                    & \url{https://github.com/Shen-Lab/GraphCL_Automated}                                         \\
GCC~\shortcite{gcc}            & \url{https://github.com/THUDM/GCC}                                          \\
InfoGCL~\shortcite{infogcl}                 & --                                               \\ 
SimGRACE~\shortcite{simGrace} & \url{https://github.com/junxia97/simgrace} \\
\midrule
GraphMAE~\shortcite{graphmae} & \url{https://github.com/THUDM/GraphMAE}\\
S2GAE~\shortcite{s2gae} & \url{https://github.com/qiaoyu-tan/S2GAE}\\

\bottomrule
\end{tabular}%
}
\caption{Baselines and their URLs in \textbf{unsupervised} representation learning.}
\label{tab:baselines}
\end{table}
\section{Introduction of Datasets and Baselines}
\label{sec:intro-datasets}

\begin{table*}[!t]
    \setlength\tabcolsep{6pt} % default value: 6pt
    \centering
    \footnotesize
    \renewcommand\arraystretch{1.4} % 行间距
    \setlength\tabcolsep{6pt} % 列间距
    \begin{tabular}{c|ccccccccc}\toprule
    &\textbf{ZINC} &\textbf{BBBP} &\textbf{Tox21} &\textbf{ToxCast} &\textbf{SIDER}  &\textbf{ClinTox} &\textbf{MUV} &\textbf{HIV} &\textbf{BACE}
    % &\textbf{Predict level} &\textbf{Predict task} &\textbf{Metric} 
    \\\midrule
    % \multirow{7}{*}{\parbox{3cm}{\centering\textbf{Pre-train GNNs}\\\cite{pretrain-gnn}}}
    \# Graphs &2,000,000 &2,039 &7,831 &8,576 &1,427 &1,477 &93,087 &41,127 &1,513
    % &graph &2-class classif. & Accuracy 
    \\
    \# Binary prediction tasks &-- &1 &12 &617 &27 & 2 &17 &1 &1
    \\
    Avg. \# nodes &26.6 &24.1 &18.6 &18.8 &33.6 &26.2 &24.2 &24.5 &34.1
    \\
    % \\\midrule
    % \multirow{3}{*}{\parbox{3cm}{\centering\textbf{Pretrain-GNNs}\\\cite{pretrain-gnn}}}
    % &ZINC & 2,000,000 &23.2 &24.9 &graph &regression &MAE \\
    % &PATTERN &14,000 &118.9 &3,039.3 &inductive node &2-class classif. &Accuracy \\
    % &CLUSTER &12,000 &117.2 &2,150.9 &inductive node &6-class classif. &Accuracy\\
    \bottomrule
    \end{tabular}
     \caption{Overview of the datasets used in \textbf{transfer} learning.}
    \label{tab:transfer-datasets}
\end{table*}

\subsection{Unsupervised Representation Learning}

\paragraph{Datasets.} In unsupervised representation learning, we use a total of seven real-world datasets, including MUTAG, IMDB-B, IMDB-M, PROTEINS, COLLAB, REDDIT-B, and NCI1, which vary in content domains and dataset sizes. The datasets used can be downloaded from PyTorch Geometric (PyG)~\cite{pytorch-geometric}~\footnote{\url{https://github.com/pyg-team/pytorch_geometric}}, and the dataset statistics are summarized in Table~\ref{tab:unsupervised-datasets}. 

\paragraph{Baselines.} To demonstrate the effectiveness of our proposed method in unsupervised representation learning, we compare StructMAE with the following 10 baseline models:  
  
  \textbf{1) Two Supervised Models:}

\begin{itemize}
  \item GIN~\cite{gin} is a graph neural network architecture designed for learning on graph-structured data. It employs a message passing scheme that incorporates neighborhood aggregation and graph isomorphism testing.
  \item DiffPool~\cite{diffpool} is a graph neural network architecture that addresses the flat nature of current GNN methods by introducing a differentiable graph pooling module, enabling hierarchical representations of graphs. 
\end{itemize}

 \textbf{2) Six Contrastive Models:}
 
 \begin{itemize}
  \item InfoGraph~\cite{infoGraph:} is a method for learning graph-level representations by maximizing mutual information between graph-level and substructure representations. 
  \item GraphCL~\cite{graphCL} is a framework for learning
unsupervised representations of graph data with various graph augmentations.
  \item JOAO~\cite{joao} is a unified bi-level optimization framework that could automatically selects augmentations, addressing the limitations of manual augmentation selection in GraphCL.
  \item GCC~\cite{gcc} is a self-supervised pre-training framework for graph neural networks, designed to capture universal network topological properties across multiple datasets.
  \item InfoGCL~\cite{infogcl} is an information-aware graph contrastive learning framework following the Information Bottleneck principle to minimize information loss during graph representation learning.
  \item SimGRACE~\cite{simGrace} is a novel framework for graph contrastive learning using the original graph and a perturbed version as inputs to two correlated encoders.
\end{itemize}

\textbf{3) Two Generative models:}

 \begin{itemize}
  \item GraphMAE~\cite{graphmae} is a masked graph autoencoder designed to address issues negatively impacting the development of graph autoencoder (GAE)~\cite{gae} in generative self-supervised learning on graphs. 
  \item  S2GAE~\cite{s2gae} jointly reconstruct the masked edges and node degrees.
  % \item DET~\cite{det} adopts two graph transformer layers.
  % \item NAGphormer~\cite{nagphormer} adopts one graph transformer layer with 8 attention heads.
  % \item ANS-GT~\cite{ans-gt} adopts 4 graph transformer layers with 8 attention heads.
  \end{itemize}

For most baselines, we refer to their implementations provided by the original paper. If unavailable, we utilize the implementation provided by PyG~\cite{pytorch-geometric}. The links of codes are included in Table~\ref{tab:baselines}.

% \textbf{4)} The architecture of our StructMAE is specified as follows:

% \begin{itemize}
%   \item StructMAE utilize a structure-based score to access the importance of each node. Two ways of scoring are proposed: predefined and learnable.
%     \item Guided by the score, StructMAE adopts an easy-to-hard masking strategy that gradually increases the difficulty of the self-supervised reconstruction task.
%   \item Predefined scoring is based on PageRank\cite{pagerank}, and learnable scoring is based on Graph Neural Network (GNN) and Multi-Layer Perceptron (MLP).
% \end{itemize}

\begin{table}[!t]
\centering
\renewcommand\arraystretch{1.4} % 行间距
\setlength\tabcolsep{2.5pt} % 列间距
\resizebox{0.48\textwidth}{!}{%
\begin{tabular}{@{}c|c@{}}
\toprule
\textbf{ Models }                & \textbf{ Code Links}                                                                                                 \\ \midrule
Infomax~\shortcite{pretrain-gnn} & \url{hhttps://github.com/snap-stanford/pretrain-gnns}                               \\
AttrMasking~\shortcite{pretrain-gnn} & \url{hhttps://github.com/snap-stanford/pretrain-gnns}     \\
ContextPred~\shortcite{pretrain-gnn} & \url{hhttps://github.com/snap-stanford/pretrain-gnns}     \\
% Graphormer~\shortcite{graphormer-v1}             & \url{https://github.com/Microsoft/Graphormer}                                             \\  
\midrule
GraphCL~\shortcite{graphCL}                  & \url{https://github.com/Shen-Lab/GraphCL} \\
JOAO~\shortcite{joao}                    & \url{https://github.com/Shen-Lab/GraphCL_Automated}                                         \\
GraphLOG~\shortcite{graphlog}          & \url{https://github.com/DeepGraphLearning/GraphLoG}                                          \\
RGCL~\shortcite{rgcl}                 &  \url{https://github.com/lsh0520/rgcl}                                          \\ 
\midrule
GraphMAE~\shortcite{graphmae} & \url{https://github.com/THUDM/GraphMAE}\\
GraphMAE2~\shortcite{graphmae2} & \url{https://github.com/thudm/graphmae2}\\
Mole-BERT~\shortcite{molebert} & \url{https://github.com/junxia97/mole-bert}\\

\bottomrule
\end{tabular}%
}
\caption{Baselines and their URLs in \textbf{transfer }learning.}
\label{tab:baselines-transfer}
\end{table}

% 7 general GCN-based models, 5 models designed for or commonly used in heterophilic settings, and 8 Transformer-based models for graphs. 
% For all GCN-based and heterophily-based methods, we For Graph Transformer baselines, we refer to their implementations provided by the original paper, and the links of 
\subsection{Transfer Learning}
\paragraph{Datasets.} In transfer learning, our model is initially pre-trained in two million unlabeled molecules sampled from the ZINC15 ~\cite{zinc15} and then finetuned in eight classification benchmark datasets contained in MoleculeNet \cite{moleculenet}, including BBBP, Tox21, ToxCast, SIDER, ClinTox, MUV, HIV, and BACE. The datasets used can be downloaded from Pretrain-GNNs~\cite{pretrain-gnn}~\footnote{\url{https://github.com/snap-stanford/pretrain-gnns}}, and the dataset statistics are summarized in Table~\ref{tab:transfer-datasets}. 
\vspace{3pt} \\
\noindent \textbf{ZINC} is a publicly available dataset developed for virtual screening, ligand discovery, pharamcophore screens, and benchmarking. ZINC15 is a new version published in 2015, containing over 120 million purchasable ``drug-like'' compounds represented as graphs. The task related to this dataset involves conducting regression analysis. 
\vspace{3pt} \\
\noindent \textbf{MoleculeNet} is a benchmark dataset specifically designed for evaluating machine learning methods on molecular properties. This comprehensive collection currently encompasses over 700,000 compounds tested across various properties. The tasks associated with this dataset can be evaluated through ROC-AUC, AUC-PRC, RMSE, and MAE scores.

\paragraph{Baselines.} To demonstrate the effectiveness of our proposed method in transfer learning, we compare StructMAE with the following 10 baseline models: 

  \textbf{1) Three Unsupervised models:}

\begin{itemize}
  \item Infomax, AttrMasking and ContextPred are three unsupervised methods from Pretrain-GNNs~\cite{pretrain-gnn}. It proposes various self-supervised methods for pre-training GNNs to address distributional differences and scarcity of task-specific labels, achieving great performance for molecular property prediction and protein function prediction.
\end{itemize}

 \textbf{2) Four Contrastive models:}
 
 \begin{itemize}
  \item GraphCL~\cite{graphCL} is a framework for learning
unsupervised representations of graph data with various graph augmentations.
  \item JOAO~\cite{joao} is a unified bi-level optimization framework that could automatically selects augmentations, addressing the limitations of manual augmentation selection in GraphCL.
  \item GraphLOG~\cite{graphlog} is a suite for studying the logical generalization capabilities of GNNs. 
  \item RGCL~\cite{rgcl} is a framework integrating invariant rationale discovery and rationale-aware pre-training.
\end{itemize}

\textbf{3) Three Generative models:}

 \begin{itemize}
  \item GraphMAE~\cite{graphmae} is a masked graph autoencoder designed to address issues negatively impacting the development of GAE~\cite{gae} in generative self-supervised learning on graphs. 
  \item GraphMAE2~\cite{graphmae2} extends GraphMAE by employing strategies such as multi-view decoding and node sampling. 
  \item Mole-BERT~\cite{molebert} represents a specialized pre-training framework designed for GNNs with a focus on molecular applications. At the heart of Mole-BERT is the innovative use of a variant of VQ-VAE, which is adeptly employed for the context-aware encoding of atom attributes.

  % \item  S2GAE~\cite{s2gae} adopts three graph transformer layers with hidden size as 64 and two attention heads.
  % \item DET~\cite{det} adopts two graph transformer layers.
  % \item NAGphormer~\cite{nagphormer} adopts one graph transformer layer with 8 attention heads.
  % \item ANS-GT~\cite{ans-gt} adopts 4 graph transformer layers with 8 attention heads.
  \end{itemize}

For most baselines, we refer to their implementations provided by the original paper. If unavailable, we utilize the implementation provided by PyG~\cite{pytorch-geometric}. The links of codes are included in Table~\ref{tab:baselines-transfer}.

% \textbf{4)} The architecture of our StructMAE is specified as follows:

% \begin{itemize}
%   \item StructMAE utilize a structure-based score to access the importance of each node. Two ways of scoring are proposed: predefined and learnable.
%     \item Guided by the score, StructMAE adopts an easy-to-hard masking strategy that gradually increases the difficulty of the self-supervised reconstruction task.
%   \item Predefined scoring is based on PageRank\cite{pagerank}, and learnable scoring is based on Graph Neural Network (GNN) and Multi-Layer Perceptron (MLP).
% \end{itemize}

% \usepackage{booktabs}
% \usepackage{graphicx}

\end{document}